\documentclass[conference]{IEEEtran}
\IEEEoverridecommandlockouts
\usepackage{graphicx}
\usepackage{multirow}
\usepackage{booktabs}
\usepackage[table]{xcolor}
\usepackage{cite}
\usepackage{amsmath,amssymb,amsfonts}
\usepackage{algorithmic}
\usepackage{graphicx}
\usepackage{textcomp}
\usepackage{xcolor}
\def\BibTeX{{\rm B\kern-.05em{\sc i\kern-.025em b}\kern-.08em
    T\kern-.1667em\lower.7ex\hbox{E}\kern-.125emX}}
\begin{document}

\title{RGBT Tracking via All-layer Multimodal Interactions with Progressive Fusion Mamba
}

\author{
	\IEEEauthorblockN{
		Andong Lu\IEEEauthorrefmark{1}, 
		Wanyu Wang\IEEEauthorrefmark{2}, 
		Chenglong Li\IEEEauthorrefmark{2}, 
		Jin Tang\IEEEauthorrefmark{1} 
		and Bin Luo\IEEEauthorrefmark{1}} 
	\IEEEauthorblockA{\IEEEauthorrefmark{1}School of Computer Science and Technology, Anhui University}
	\IEEEauthorblockA{\IEEEauthorrefmark{2}School of Artificial Intelligence, Anhui University\\  adlu\_ah@foxmail.com}
}


\maketitle

\begin{abstract}

%
Existing RGBT tracking methods often design various interaction models to perform cross-modal fusion of each layer, but can not execute the feature interactions among all layers, which plays a critical role in robust multimodal representation, due to large computational burden.
To address this issue, this paper presents a novel All-layer multimodal Interaction Network, named AINet, which performs efficient and effective feature interactions of all modalities and layers in a progressive fusion Mamba, for robust RGBT tracking. 
Even though modality features in different layers are known to contain different cues, it is always challenging to build multimodal interactions in each layer due to struggling in balancing interaction capabilities and efficiency.
Meanwhile, considering that the feature discrepancy between RGB and thermal modalities reflects their complementary information to some extent, we design a Difference-based Fusion Mamba (DFM) to achieve enhanced fusion of different modalities with linear complexity.
When interacting with features from all layers, a huge number of token sequences (3840 tokens in this work) are involved and the computational burden is thus large. To handle this problem, we design an Order-dynamic Fusion Mamba (OFM) to execute efficient and effective feature interactions of all layers by dynamically adjusting the scan order of different layers in Mamba.
Extensive experiments on four public RGBT tracking datasets show that AINet achieves leading performance against existing state-of-the-art methods.
\end{abstract}

\section{Introduction}

RGBT tracking aims to leverage the complementary information of visible light (RGB) and thermal infrared (TIR) images to predict the location and size of an object. By combining the penetration capability and nighttime sensitivity of TIR images with the rich color and texture of RGB images under favorable lighting conditions, RGBT tracking has attracted significant research attention. Numerous innovative works~\cite{tbsi,QAT2023,TATrack,BAT2024,ckd} have been proposed to improve the robustness and accuracy of RGBT tracking.

\begin{figure}[t]
    \centering
    \includegraphics[width=0.96\linewidth]{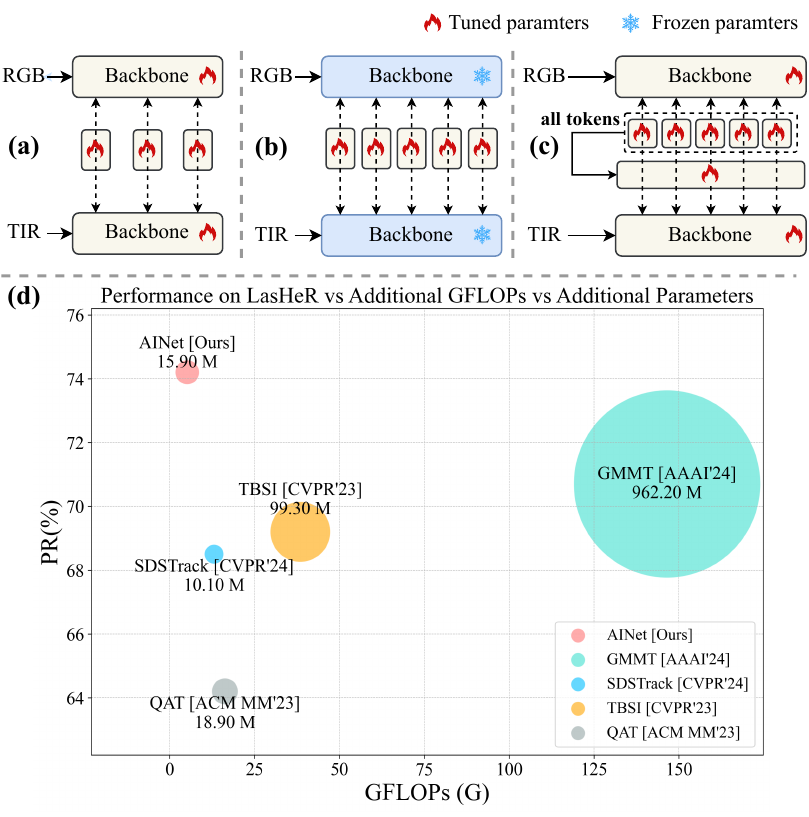}
     \caption{Comparison with existing RGBT tracking methods. (a) Interactions between specific layers, with joint fine-tuning of the entire backbone. (b) Interactions between all corresponding layers, with the pre-trained backbone being frozen. (c) Interactions between all corresponding layers, and interactions among all layers, with joint fine-tuning with the backbone. (d) Performance comparison on LasHeR, and comparison of additional parameters and GFLOPs.}
    \label{motivation}
\end{figure}


As a multimodal vision task, most RGBT tracking studies focus on modality interaction modules, and these approaches can be broadly classified into two categories.
One category involves building complex interaction modules~\cite{tbsi,fan2024querytrack,gmmt} with powerful representation to achieve inter-modal interactions.
For example, ~\cite{tbsi} employ six cross-attention blocks and extra fused template features for inter-modal feature interactions. Similarly, ~\cite{fan2024querytrack} stack
multiple Swin-Transformer~\cite{liu2021swin} blocks to perform self-attention between two modality features. 
However, due to the significant computational burden, these methods can only interact at a limited number of specific layers, as shown in Fig.~\ref{motivation}(a).
Another category involves designing lightweight interaction modules and adopting the strategy of interacting between all corresponding layers to achieve inter-modal interactions, as illustrated in Fig.~\ref{motivation} (b).
For instance, ~\cite{BAT2024} deploy lightweight adapters in all layers to perform bi-directional inter-modal interactions. Similar interaction strategies are also demonstrated in \cite{SDSTrack,vipt}.  
However, these methods have restricted interaction capabilities and representation capacity due to their extremely low parameter count.
%
As a result, existing methods struggle to balance interaction capability and efficiency when constructing multimodal interactions between all corresponding layers.

In addition, features from different layers show significant complementarity: low-level features provide detailed texture information, while high-level features capture abstract and semantic content. 
Nevertheless, existing methods adopt CNNs or Transformer networks, where the small receptive field of CNNs hinders the modeling of global information between modalities, and the Transformer is hard to interact with information from multi-layer features due to the $O(N^2)$ computational complexity. Hence, these architectural limitations prevent previous RGBT tracking methods from achieving comprehensive interaction of all layer information, which is essential for robust RGBT tracking.

To address these issues, we propose a novel All-layer multimodal Interaction Network named AINet, as shown in Fig.~\ref{motivation} (c), for robust RGBT tracking.
In particular, modality feature interaction aims to leverage the mutual enhancement and complementarity of two modalities, while complementary information reflected in their differences.
Thus, we design a difference-based fusion Mamba (DFM) with linear complexity, which not only can model modality differences to enhance each modality feature, but also can be efficiently applied to each layer. 
When performing feature interactions across all layers, handling a large number of token sequences (3840 tokens) results in a significant computational burden for existing networks. Although the standard Mamba network is efficient, it is prone to forgetting early token information due to the inherent properties of causal models. To mitigate this issue, we design an Order-dynamic Fusion Mamba (OFM) module, which dynamically adjusts the scan order of different layers based on the input to alleviate layer information forgetting.
Extensive experiments on four public RGBT tracking benchmarks show that AINet significantly outperforms existing state-of-the-art methods in both performance and efficiency, as depicted in Fig.~\ref{motivation} (d). Our main contributions are summarized as follows:
\begin{itemize}
   \item We propose a novel all-layer multimodal interaction network for RGBT tracking. It conducts multimodal interaction of each layer and all layer interaction in a progressive fusion Mamba. To the best of our knowledge, we are the first to introduce the Mamba network in RGBT tracking.
   
  \item We design a difference-based fusion Mamba, which achieves inter-modal enhanced fusion by modelling inter-modal differences to capture complementary information, and efficiently applies it to each layer. 

  \item We design an order-dynamic fusion Mamba, which implements all-layer feature interaction with an input-aware dynamic scanning scheme to mitigate information forgetting of early input tokens.

  
  \item Extensive experiments on four RGBT benchmarks demonstrate that AINet achieves new state-of-the-art results while maintaining a controllable number of parameters and computational load.

\end{itemize}

\section{Related Work}

\subsection{RGBT Tracking}
In recent years, the field of RGBT tracking has made significant progress, with many impressive works emerging.
Existing methods can be roughly divided into two categories based on their feature fusion strategies. One category involves late fusion, which takes place after the main feature extraction backbone. For instance, \cite{mfdimp} embed the multimodal feature concatenation process into the framework of a strong tracker DiMP~\cite{dimp} for RGBT tracking. \cite{peng2023siamese} utilize two-stream convolutional network with increasing coupling filters to extract both common and individual features, and ultimately achieves fusion through a simple channel concatenation. The second category involves modality interaction during the feature extraction phase. For example, \cite{tbsi} build a template-bridged cross-attention module between the RGB and TIR search areas to promote thorough modality fusion. \cite{BAT2024} propose a bi-directional adapter to perceive the dominant modality changes in dynamic scenes and adaptively fuse multimodal information. However, existing methods are limited by computational costs and cannot utilize information from all layers. Our method can fully leverage all-layer multimodal information while keeping a small number of parameters and low computational resource consumption.

\subsection{Vision State Space Model}
State space models (SSMs)~\cite{ssm1,ssm2,ssm3} derived from classical control theory connects the input and output sequences through hidden states. Recently, Mamba~\cite{mamba} has been widely applied in various fields due to its selective mechanism and efficient hardware acceleration design. \cite{vim} expand to visual tasks for the first time by bidirectional sequence modeling. \cite{mambair} propose a residual state space block, using convolution and channel attention to enhance the performance of Vanilla Mamba, and achieved success in the field of image restoration. \cite{motionmamba} integrate a selective scanning mechanism into the motion generation task, constructing HTM and BSM modules to handle temporal motion data and bidirectionally capture the channel-wise flow of hidden information within the latent pose. However, the application of Mamba in RGBT tracking has not yet been sufficiently explored. In this work, we leverage Mamba for modality enhancement and all-layer fusion, exploring the potential of Mamba in RGBT tracking.

\begin{figure*}[t]
    \centering
    \centerline{\includegraphics[width=0.95\textwidth]{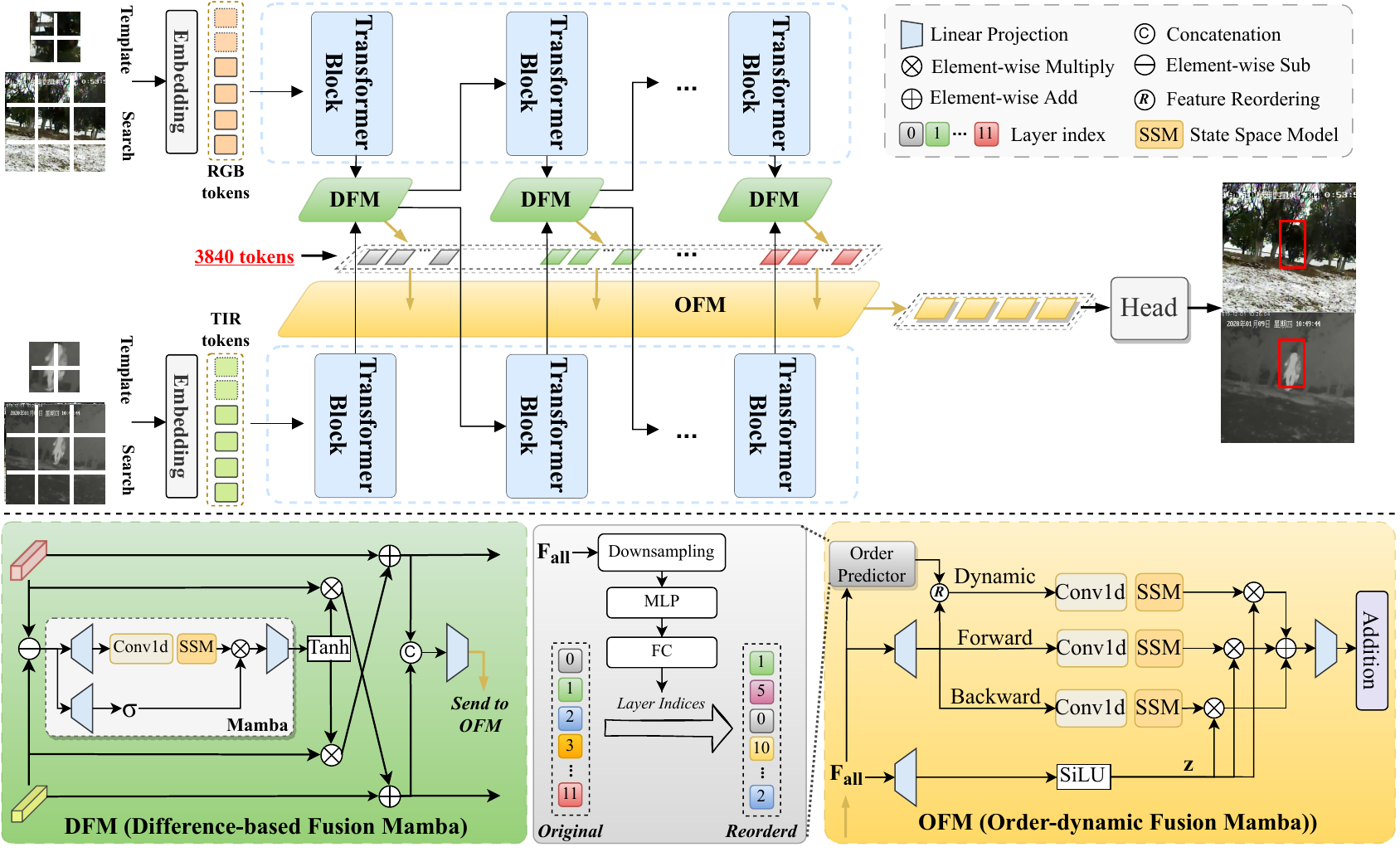}}
    \caption{
    The overall architecture of our proposed AINet. Firstly, RGB and TIR images are embedded as tokens and fed into Transformer blocks for joint feature extraction and relationship modeling between search and template images. Following each block, the tokens from both modalities are processed by the DFM for difference information enhancement and then returned to the backbone. Meanwhile, the fusion features at each layer are cascaded and fed into the OFM for all-layer interaction. Finally, the output features from the OFM are sent to the tracking head for target localization.
     }
    \label{architecture}
\end{figure*}

\section{Methodology}

In this paper, we propose a novel All-layer multimodal Interaction Network, named AINet, for RGBT tracking, which performs efficient and effective feature interactions of all modalities and layers in a progressive fusion Mamba. In particular, AINet achieves multimodal interactions at each layer by the designing difference-based fusion Mamba, and it employs an order-dynamic fusion Mamba to establish all-layer interactions. 
Next, we first review the mamba, then introduce the overall architecture of AINet, and finally, we describe in detail the two fusion Mamba architectures.

\begin{figure}[h]
    \centering
    \includegraphics[width=1\linewidth]{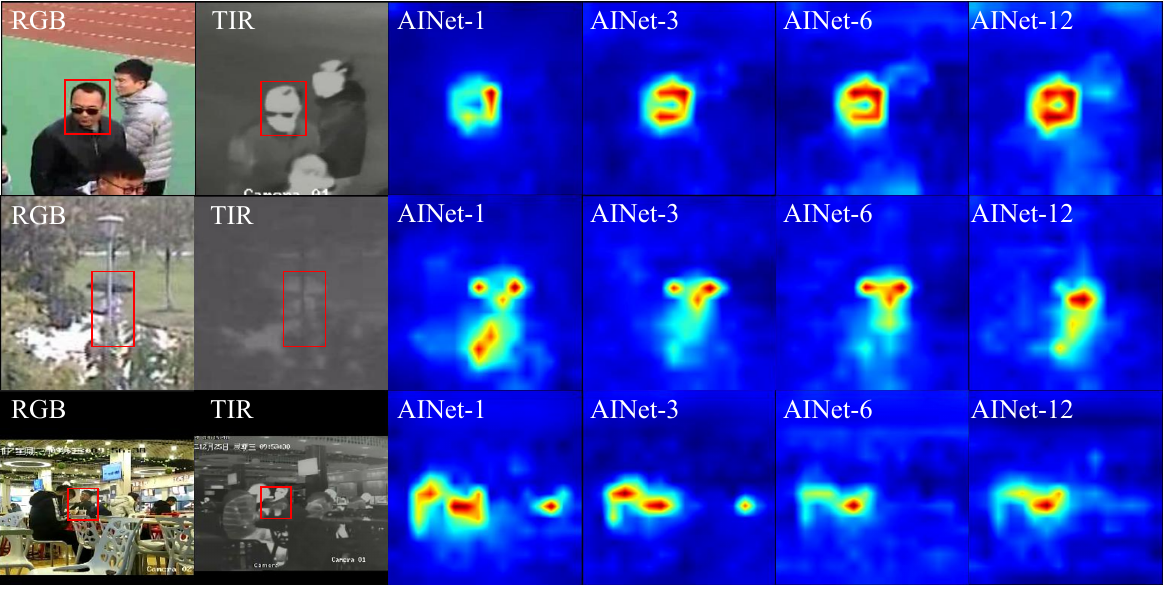}
     \caption{Illustration of fusion feature visualization with different layers applied. Here, “n” in AINet-“n” represents the number of layers applied.}
    \label{cross_feats}
\end{figure}

To visually demonstrate the necessity of applying all layer features, we perform a visual analysis of the final fusion features that incorporate different numbers of layer features, as shown in Fig.~\ref{cross_feats}. It can be observed that as the number of layers increases, the model's response to the target becomes more comprehensive and focused, validating the necessity and importance of applying all layer features.



\subsection{Preliminaries}
The state-space sequence model (SSM)~\cite{ssm1} and Mamba~\cite{mamba} are inspired by continuous linear systems, where a one-dimensional function or sequence, denoted as $x(t)\in \mathbb{R}$, is mapped to $x(t)\in \mathbb{R}$ through a hidden states $h(t)\in \mathbb{R}^{N}$. The models can be represented by linear ordinary differential equations (ODEs) as follows:
\begin{equation}
h^{'}(t)=\boldsymbol{A}h(t)+\boldsymbol{B}x(t),\ y(t)=\boldsymbol{C}h(t),\label{con:eq1}
\end{equation}
where $N$ is the state size, $\boldsymbol{A}\in \mathbb{R}^{N\times{N}}$, $\boldsymbol{B}\in \mathbb{R}^{N\times{1}}$, $\boldsymbol{C}\in \mathbb{R}^{1\times{N}}$. 
Specifically, $\Delta$ denotes the timescale parameter to transform the continuous parameters A, B to discrete parameters A, B. The commonly used method for discretization is the zero-order hold (ZOH) rule, which is defined as follows:
\begin{equation}
\overline{\boldsymbol A} = \exp(\Delta \boldsymbol A),\
\overline{\boldsymbol B} = (\Delta \boldsymbol A)^{-1}(\exp(\Delta \boldsymbol A)-\boldsymbol I)\cdot \Delta \boldsymbol B.
\end{equation}

Then, the discretized version of Eq.~\ref{con:eq1} with step size $\Delta$ can be rewritten as:
\begin{equation}
h_t=\overline{\boldsymbol A}h_{t-1}+\overline{\boldsymbol B}x_t,\ y_t=\boldsymbol{C}h_t.
\end{equation}

Finally, the models compute output $\boldsymbol{\mathrm {y}}$ through a global convolution within a structured convolutional kernel $\boldsymbol{\overline K}$.
\begin{equation}
\boldsymbol{\overline K=(C \overline B,C \overline {AB},\dots,
C {\overline A}^{M-1}\overline B)},\ \boldsymbol{\mathrm{y}} =\boldsymbol{\mathrm {x}}*\boldsymbol{\overline K},
\end{equation}
where $\boldsymbol M$ denotes the sequence length of $\boldsymbol{\mathrm {x}}$. In contrast to traditional SSMs, Mamba introduces the Selective Scanning Mechanism (S6), which allows selective focus on what is important in long contexts. In this work, we extend the Mamba module to support all-layer multimodal interactions.

\subsection{Overall Architecture}

Inspired by the success of ViT~\cite{vit} in tracking tasks, we follow the OSTrack~\cite{ostrack} framework and extend its backbone to adapt to multimodal input.
As shown in Fig.~\ref{architecture}, our AINet incorporates a two-stream encoder structure, sharing the same parameters for both the RGB and TIR modalities. Additionally, it includes a set of difference-based fusion Mamba (DFM), an order-dynamic fusion Mamba (OFM), and a prediction head.
Specifically, AINet first processes the search and template frames of the given RGB and TIR modalities through the patch and position embedding layers, obtaining the initial token pairs for each modality. The search and template tokens from each modality are then concatenated along the token dimension to form RGB tokens $x_{0}^{rgb}$ and TIR tokens $x_{0}^{tir}$, which are fed into the ViT blocks for feature extraction and joint relationship modeling.
Since the DFM is embedded in each layer to facilitate modality interactions, for the $i$-th layer block, it learns to integrate enhanced modality features from the output of the previous DFM, as described below.
\begin{equation}
(\hat{x}_{i}^{rgb},\hat{x}_{i}^{tir},x_{i}^{dfm})=\mathcal{F}^{DFM}_i(x_{i}^{rgb},x_{i}^{tir}),i\in \left[1,N\right],
\end{equation}
where $N$ refers to the total number of blocks. $\hat{x}_{i}^{rgb}$ and $\hat{x}_{i}^{tir}$ represent the enhanced RGB and TIR modality features, respectively. $\mathcal{F}^{DFM}_i$ and $x_{i}^{dfm}$ denote the DFM and the fused features, respectively, at the $i$-th layer.
%
When all blocks are executed, $x_{i}^{dfm}$ from all layers are concatenated along the token dimension and fed into the OFM, represented as $\mathcal{F}^{OFM}_i$, for all layers to interact. Then, the prediction head $\mathcal{H}$ predicts the tracking result $\mathcal{B}$ based on the output of the OFM.
\begin{equation}
\mathcal{B}=\mathcal{H}(\mathcal{F}^{OFM}[x_{1}^{dfm},x_{2}^{dfm},\dots{},x_{N}^F]),
\end{equation}
where $[\cdot]$ refers to concatenation along the token dimension.

\subsection{Difference-based Fusion Mamba (DFM)}

Visible and thermal infrared modalities capture complementary object properties due to different imaging principles, which implies that the differences between modalities often contain complementary information. Nevertheless, current advanced approaches mainly adopt Transformer networks with long-range modeling capabilities to directly interact the features of the modalities, which ignores the explicit utilization of modality differences and limits multimodal interactions at each layer due to high secondary computational overhead. To this end, we design a difference-based fusion Mamba (DFM) with linear complexity, as depicted in Fig.~\ref{architecture} (b), which can be employed to model modality differences at each layer to enhance modal representation. 
To capture inter-modal differences and enrich feature learning, DFM employs the principles of differential amplifier circuits, which suppresses common-mode signals and amplifies differential ones. 

In particular, we obtain the modality difference feature $x_{i}^{d}$ by subtracting between modal features in the same layer. Since the difference feature contains both useful information and noise, $x_{i}^{d}$ is then fed to the Mamba to suppress noise while enhancing useful information. Next, $x_{i}^{d}$, processed by the activation function, is multiplied element-wise with $x_{i}^{rgb}$ and $x_{i}^{tir}$ to obtain the difference compensation features for each modality. Finally, these compensated features are added back to $x_{i}^{rgb}$ and $x_{i}^{tir}$ to obtain the enhanced features $\hat{x}_{i}^{rgb}$ and $\hat{x}_{i}^{tir}$. The following equation summarizes the process:
\begin{equation}
\begin{aligned}
& \hat{x} _{i}^{rgb}=  x _{i}^{rgb} + x _{i}^{tir}\times\tau (Mamba(x_{i}^{d})),\\
& \hat{x} _{i}^{tir}=  x _{i}^{tir} + x _{i}^{rgb}\times\tau (Mamba(x_{i}^{d})),
\end{aligned}
\end{equation}
where $\tau$ denotes the $tanh$ function. Subsequently, the enhanced modality features are processed to obtain the fused feature of the current layer:
\begin{equation}
x _i^{dfm}=\tau(LN([x _i^{rgb},x _i^{tir}]\cdot W_i)),
\end{equation}
where $W_i\in \mathbb{R}^{2C\times C}$ is a linear layer with reduced channel dimension, and $LN$ represents layer normalization.

\subsection{Order-dynamic Fusion Mamba (OFM)}
The strong complementarity between different feature layers in deep networks has been proven in many visual tasks~\cite{fpn,pang2020multi,liu2023improving}. However, no current method applies all feature layers to RGBT tracking, primarily because existing methods use Transformers for feature interactions.
To this end, we design an Order-dynamic Fusion Mamba (OFM) to efficiently and effectively interact with features from all layers by dynamically adjusting the scan order of different layers in Mamba.

In particular, We first concatenate the output features $x^{dfm}_i$ of each DFM layer along the token dimension to form a long token sequence $F_{all}$ containing features from all layers. We then input $F_{all}$ to the OFM and perform the following forward and backward scanning modeling process:
\begin{equation}
\begin{aligned}
& F_{all}^{forward} = \mathcal{SSM}([x_{1}^{dfm},x_{2}^{dfm},\dots{},x_{N}^{dfm}],W_c), \\
& F_{all}^{backward} = \mathcal{SSM}([x_{N}^{dfm},x_{N-1}^{dfm},\dots{},x_{1}^{dfm}],W_c),
\end{aligned}
\end{equation}
where $W_c$ represents a 1D convolution layer and $\mathcal{SSM}$ denotes the selective scanning model. However, only performing forward and backward scans can potentially ignore the first and last layer tokens. Therefore, we propose an order-dynamic scanning scheme that allows the scanning process to start and end at any layer. This innovative design enables OFM to rank the importance of different layer features based on the input data.

In the order-dynamic scanning modeling, $F_{all}$ is first downsampled ($\mathcal{D}$) to the specified dimension and then fed into a multi-layer perceptron ($\mathcal{MLP}$) and a fully connected layer ($\mathcal{FC}$) to predict an index covering the scanning order of all layers. The process is expressed as follows:
\begin{equation}
\begin{aligned}
&index=\mathcal{FC}(\mathcal{MLP}(\mathcal{D}(F))). \\
\end{aligned}
\end{equation}

Then, the OFM reorders the long token sequence by the $index$. Thus, the dynamic ordering scan modeling can be formulated as follows:
\begin{equation}
\begin{aligned}
& x^{order}_{all} = \{[x_{1}^{dfm},x_{2}^{dfm},\dots{},x_{N}^{dfm}], index\},\\
& F_{all}^{dynamic} = \mathcal{SSM}(x^{order}_{all},W_c),
\end{aligned}
\end{equation}
where $\{\cdot, index\}$ represents the input sequence ordered according to the given index, and $x^{order}_{all}$ denotes the ordered result. Next, a simple gating strategy in Mamba fuses the results of three-scan modeling. Finally, all layer features are aggregated by element-wise addition and fed into the tracking head.

\section{Experiment}
\begin{table*}[ht]
    \centering
     \caption{The PR, NPR, and SR scores (\%) of various trackers on five datasets. The best and second results are in \textcolor{red}{red} and \textcolor{blue}{blue} colors, respectively.}
    \setlength{\tabcolsep}{1.3mm}{
    \renewcommand\arraystretch{1.0}{
	\resizebox{\linewidth}{!}{
	\begin{tabular}{c|c|c|cc|cc|ccc|cc|c}
		\toprule
		\multirow{2}{*}{Methods} & \multirow{2}{*}{Pub. Info.} & \multirow{2}{*}{Backbone} & \multicolumn{2}{c|}{RGBT210} & \multicolumn{2}{c|}{RGBT234} & \multicolumn{3}{c|}{LasHeR} & \multicolumn{2}{c|}{VTUAV} & {FPS} \\
		& & & PR$\uparrow$ & SR$\uparrow$ & PR$\uparrow$ & SR$\uparrow$ & PR$\uparrow$ & NPR$\uparrow$ & SR$\uparrow$ & PR$\uparrow$ & SR$\uparrow$ & $\uparrow$ \\
		\hline
        mfDiMP~\cite{zhang2019multi} & ICCVW 2019 & ResNet$-$50 & 78.6 & 55.5 & $-$ & $-$ & 44.7 & 39.5 & 34.3 & 67.3 & 55.4  & 10.3 \\
        CAT~\cite{2020CAT} & ECCV 2020 & VGG$-$M & 79.2 & 53.3 & 80.4 & 56.1 & 45.0 & 39.5 & 31.4 & $-$ & $-$ & 20\\
        ADRNet~\cite{ADRNet2021} & IJCV 2021 & VGG-M & $-$ & $-$ & 80.7 & 57.0 & $-$ & $-$ & $-$ & 62.2 & 46.6 & 25\\
        APFNet~\cite{APFNet2022} & AAAI 2022 & VGG$-$M & $-$ & $-$ & 82.7 & 57.9 & 50.0 & 43.9 & 36.2 & $-$ & $-$ & 1.3 \\
        DMCNet~\cite{dmcnet} & TNNLS 2022 & VGG$-$M & 79.7 & 55.5 & 83.9 & 59.3 & 49.0 & 43.1 & 35.5 & $-$ & $-$ & 2.3 \\
        ProTrack~\cite{ProRGBTTrack} & ACM MM 2022 & ViT$-$B & $-$ & $-$ & 78.6 & 58.7 & 50.9 & $-$ & 42.1 & $-$ & $-$ & 30 \\
        HMFT~\cite{Zhang_CVPR22_VTUAV} & CVPR 2022 & ResNet$-$50 & 78.6 & 53.5 & 78.8 & 56.8 & $-$ & $-$ & $-$ & 75.8 & 62.7 & 30.2 \\
        MFG~\cite{wang2022mfgnet} & TMM 2022 & ResNet$-$18 & 74.9 & 46.7 & 75.8 & 51.5 & $-$ & $-$ & $-$ & $-$ & $-$ & $-$ \\
        DFNet~\cite{peng2022dynamic} & TITS 2022 & VGG$-$M & $-$ & $-$ & 77.2 & 51.3 & $-$ & $-$ & $-$ & $-$ & $-$ & $-$ \\
        DRGCNet~\cite{mei2023differential} & IEEE SENS J 2023 & VGG$-$M & $-$ & $-$ & 82.5 & 58.1 & 48.3 & 42.3 & 33.8 & $-$ & $-$ & 4.9 \\
        CMD~\cite{zhang2023efficient} & CVPR 2023 & ResNet$-$50 & $-$ & $-$ & 82.4 & 58.4 & 59.0 & 54.6 & 46.4 & $-$ & $-$ & 30 \\
        ViPT~\cite{vipt} & CVPR 2023 & ViT$-$B & $-$ & $-$ & 83.5 & 61.7 & 65.1 & $-$ & 52.5 & $-$ & $-$ & $-$ \\
        TBSI~\cite{tbsi} & CVPR 2023 & ViT$-$B & 85.3 & 62.5 & 87.1 & 63.7 & 69.2 & 65.7 & 55.6 & $-$ & $-$ & 36.2 \\
        QAT~\cite{QAT2023} & ACM MM 2023 & ResNet$-$50 & \textcolor{blue}{86.8} & 61.9 & 88.4 & 64.4 & 64.2 & 59.6 & 50.1 & 80.1 & 66.7 & 22 \\ \hline
        
        TATrack~\cite{TATrack} & AAAI 2024 & ViT$-$B & 85.3 & 61.8 & 87.2 & 64.4 & 70.2 & 66.7 & 56.1 & $-$ & $-$ & 26.1 \\
        BAT~\cite{BAT2024} & AAAI 2024 & ViT$-$B & $-$ & $-$ & 86.8 & 64.1 & 70.2 & $-$ & 56.3 & $-$ & $-$ & $-$ \\
        GMMT~\cite{gmmt} & AAAI 2024 & ViT$-$B & $-$ & $-$ & 87.9 & 64.7 & 70.7 & 67.0 & 56.6 & $-$ & $-$ & $-$ \\
        OneTracker~\cite{OneTracker} & CVPR 2024 & ViT$-$B & $-$ & $-$ & 85.7 & {64.2} & 67.2 & $-$ & 53.8 & $-$ & $-$ & $-$ \\
        {Un-Track}~\cite{Un-Track} & CVPR 2024 & ViT$-$B & $-$ & $-$ & 84.2 & 62.5 & 66.7 & $-$ & 53.6 & $-$ & $-$ & $-$ \\
        SDSTrack~\cite{SDSTrack} & CVPR 2024 & ViT$-$B & $-$ & $-$ & 84.8 & 62.5 & 66.5 & $-$ & 53.1 & $-$ & $-$ & 20.9 \\
  	\hline
	AINet (256$\times$256) &  $-$ & ViT$-$B & \textcolor{blue}{86.8} & \textcolor{blue}{64.1} & \textcolor{blue}{89.1} & \textcolor{blue}{66.8} & \textcolor{blue}{73.0} & \textcolor{blue}{69.0} & \textcolor{blue}{58.2} & \textcolor{blue}{87.1} & \textcolor{blue}{74.5} & \textcolor{red}{38.1} \\ 
    AINet (384$\times$384) &  $-$ & ViT$-$B & \textcolor{red}{87.5} & \textcolor{red}{64.8} & \textcolor{red}{89.2} & \textcolor{red}{67.3} & \textcolor{red}{74.2} & \textcolor{red}{70.1} & \textcolor{red}{59.1} & \textcolor{red}{88.0} & \textcolor{red}{75.3} & \textcolor{blue}{37.5} \\ \bottomrule
       \end{tabular}}}}

\label{overall_result}
\end{table*}

\subsection{Implementation Details}
We implement our AINet based on the PyTorch and train it on single NVIDIA RTX 4090 GPU. We follow the hyper-parameter settings of the baseline model~\cite{ostrack} for the loss function. For parameter initialization, we utilize the pretrained model provided by DropTrack~\cite{wu2023dropmae}. For each sequence in the training set, we collect training samples and apply standard data augmentation operations, including rotation, translation, and gray-scaling, following the data processing scheme of the base tracker~\cite{ostrack}. During training, we use the AdamW~\cite{AdamW} optimizer with a weight decay of $10^{-4}$, and set the batch size and learning rate to 16 and $10^{-4}$, respectively. 
The entire network is trained end-to-end over 15 epochs, with each epoch providing $6\times 10^4$ pairs of samples. We use the LasHeR training set to train our network, which is used to evaluate RGBT210~\cite{Li17rgbt210}, RGB234~\cite{li2019rgb234} and LasHeR~\cite{li2021lasher}. For the evaluation of VTUAV~\cite{Zhang_CVPR22_VTUAV}, we utilize the training set from VTUAV as the training data.

\subsection{Quantitative Comparison}

We evaluate our proposed AINet on four popular RGBT tracking benchmarks: RGBT210, RGBT234, LasHeR and VTUAV, 
and compare the performance with 20 state-of-the-art RGBT trackers. We adopt the Precision Rate (PR), Success Rate (SR), and Normalized Precision Rate (NPR) from One-Pass Evaluation (OPE) as metrics for quantitative performance measurement, which are commonly used in current RGBT tracking tasks. The effectiveness of our proposed AINet is demonstrated in Table~\ref{overall_result}, which summarizes the comparison results.

\noindent\textbf{Evaluation on RGBT210 dataset.} RGBT210 is a challenging dataset that contains 210 pairs of RGBT video sequences, totaling approximately 210K frames, and provides annotations for 12 different challenge attributes. As shown in Table~\ref{overall_result}, AINet achieves the best performance, outperforming the current state-of-the-art QAT and TATrack by 0.7\%/2.9\% and 2.2\%/3.0\% in PR/SR, respectively.

\noindent\textbf{Evaluation on RGBT234 dataset.} RGBT234 is one of the most influential and extensively evaluated RGBT tracking dataset, containing 234 pairs of aligned RGBT videos, amounting to approximately 233.4K frames. As shown in Table~\ref{overall_result}, we evaluate AINet against 20 state-of-the-art trackers on RGBT234 dataset. Compared with QAT and GMMT, the second best-performing algorithms in PR and SR respectively, our method shows a performance advantage of 0.8\%/2.9\% in PR/SR and 1.3\%/2.6\% in PR/SR respectively.


\noindent\textbf{Evaluation on LasHeR dataset.} LasHeR is the largest and most challenging RGBT tracking dataset, containing 1,224 aligned RGBT video sequences totaling approximately 734.8K frames. It also provides annotations for 19 challenge attributes, such as HI (High Illumination) and AIV (Abrupt Illumination Variation), significantly raising the dataset's challenge.

\textit{1) Overall Comparison.} As shown in Table~\ref{overall_result}, we compare our method with 17 state-of-the-art trackers using PR, NPR, and SR metrics on the LasHeR dataset. The results demonstrate that our method significantly outperforms other trackers. Specifically, compared to the most powerful tracker, GMMT, our method achieves improvements of 3.5\%/3.1\%/2.5\% in PR/NPR/SR. Compared to the unified frameworks OneTracker, Un-Track, and SDSTrack, we achieve performance improvements of 7.0\%/5.3\%, 7.5\%/5.5\%, and 7.7\%/6.0\% in PR/SR, respectively. These results fully demonstrate the superiority of our method.

\begin{figure}[t]
    \centering
    \includegraphics[width=1\linewidth]{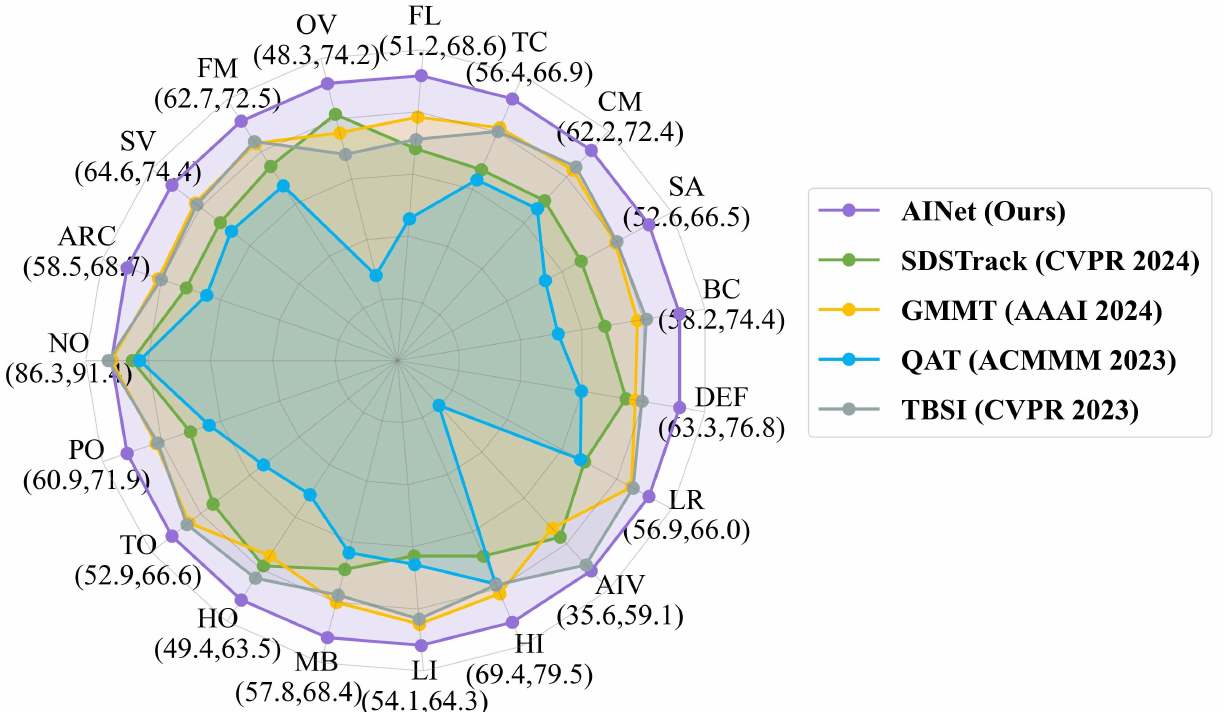}
     \caption{Precision Rate (PR) of challenge attributes on LasHeR. The axes of each attribute have been normalized.}
    \label{radar_pr}
\end{figure}

\textit{2) Challenge-based Comparison.} We also present the results of AINet against the most advanced RGBT trackers, including SDSTrack~\cite{SDSTrack}, GMMT~\cite{gmmt}, QAT~\cite{QAT2023}, and TBSI~\cite{tbsi}, on different challenge subsets. The evaluation results are shown in Fig~\ref{radar_pr}, where each corner represents the attributes of the challenge subset and the highest and lowest performance under that attribute. The challenge subsets include no occlusion (NO), partial occlusion (PO), total occlusion (TO), hyaline occlusion (HO), motion blur (MB), low illumination (LI), high illumination (HI), abrupt illumination variation (AIV), low resolution (LR), deformation (DEF), background clutter (BC), similar appearance (SA), camera moving (CM), thermal crossover (TC), frame lost (FL), out-of-view (OV), fast motion (FM), scale variation (SV), and aspect ratio change (ARC). The results show that AINet achieves the best performance in almost all subsets, especially with significant improvements in scenarios like MB, DEF, and FL, proving that AINet has great potential in various complex tracking scenarios.

\noindent\textbf{Evaluation on VTUAV dataset.} VTUAV collects RGBT data
from UAV scenarios, expanding the application of RGBT tracking. It contains 500 aligned RGBT video sequences with up to 1.7 million frames. We focus our experiments on its short-term tracking subset. As shown in Table~\ref{overall_result}, AINet achieves a PR/SR performance of 88.0\%/75.3\%, outperforming four other trackers. Compared to QAT~\cite{QAT2023}, the second-best algorithm on this dataset, AINet shows an advantage of 7.9\%/8.6\% in PR/SR. These results demonstrate the advantages of our method under the UAV perspective.


\subsection{Ablation Study}
       

\begin{table}[]
    \centering
    \caption{Quantitative comparison of different variants of our method on the LasHeR dataset. OFM$^\ast$ indicates the removal of the order-dynamic scanning scheme within the OFM.}
    \setlength{\tabcolsep}{2.5mm}
    \renewcommand{\arraystretch}{1}{
    \resizebox{\linewidth}{!}{
	\begin{tabular}{c|c|ccc}
		\toprule
		Methods & Resolution & PR$\uparrow$ & NPR$\uparrow$ & SR$\uparrow$ \\ \hline
        Baseline & 256$\times$256 & 71.1 & 67.5 & 56.9 \\
        w/ DFM & 256$\times$256& 72.1 & 68.3 & 57.4 \\
        w/ OFM & 256$\times$256& 72.2 & 68.0 & 57.3 \\
        w/ DFM OFM$^\ast$ & 256$\times$256 & 72.5 & 68.6 & 57.8 \\
        w/ DFM OFM & 256$\times$256 & 73.0 & 69.1 & 58.2 \\
        w/ DFM OFM & 384$\times$384 & \textbf{74.2} & \textbf{70.1} & \textbf{59.1} \\
        \bottomrule
    \end{tabular}}}
    \label{ab_component}
\end{table}


\noindent\textbf{Component analysis.}
In Table~\ref{ab_component}, we conduct several ablation studies on the LasHeR dataset to verify the effectiveness of key components in our AINet.

\textit{\textbf{Baseline}} denotes the removal of DFM and OFM modules from our method, while maintaining consistent training data and losses.

\textit{\textbf{w/ DFM}} indicates that each layer in the Baseline backbone network is equipped with a DFM module, achieving improvements of 1\%/0.8\%/0.5\% in PR/NPR/SR, respectively. This experiment shows that difference-based fusion Mamba is effective.

\textit{\textbf{w/ OFM}} denotes that each layer in the Baseline backbone network uses simple feature addition to obtain fused features and is equipped with our proposed OFM module, achieving improvements of 1.1\%/0.5\%/0.4\% in PR/NPR/SR, respectively. This experiment shows that order-dynamic fusion Mamba is effective.

\textit{\textbf{w/ DFM OFM}} represents applying both our proposed DFM and OFM modules in the Baseline, achieving a clear performance improvement of 1.9\%/1.6\%/1.3\% in PR/NPR/SR metrics, respectively. This experiment demonstrates the effectiveness of our proposed progressive fusion Mamba. In addition, we remove our designed order-dynamic scanning scheme in \textit{\textbf{w/ DFM OFM}} and denote it as \textit{\textbf{w/ DFM OFM$^\ast$}}. The results show a certain decrease, proving the effectiveness and necessity of order-dynamic scanning.


%
%

\definecolor{c1}{HTML}{00b894}
\begin{figure}[t]
    \centering
    \includegraphics[width=1\linewidth]{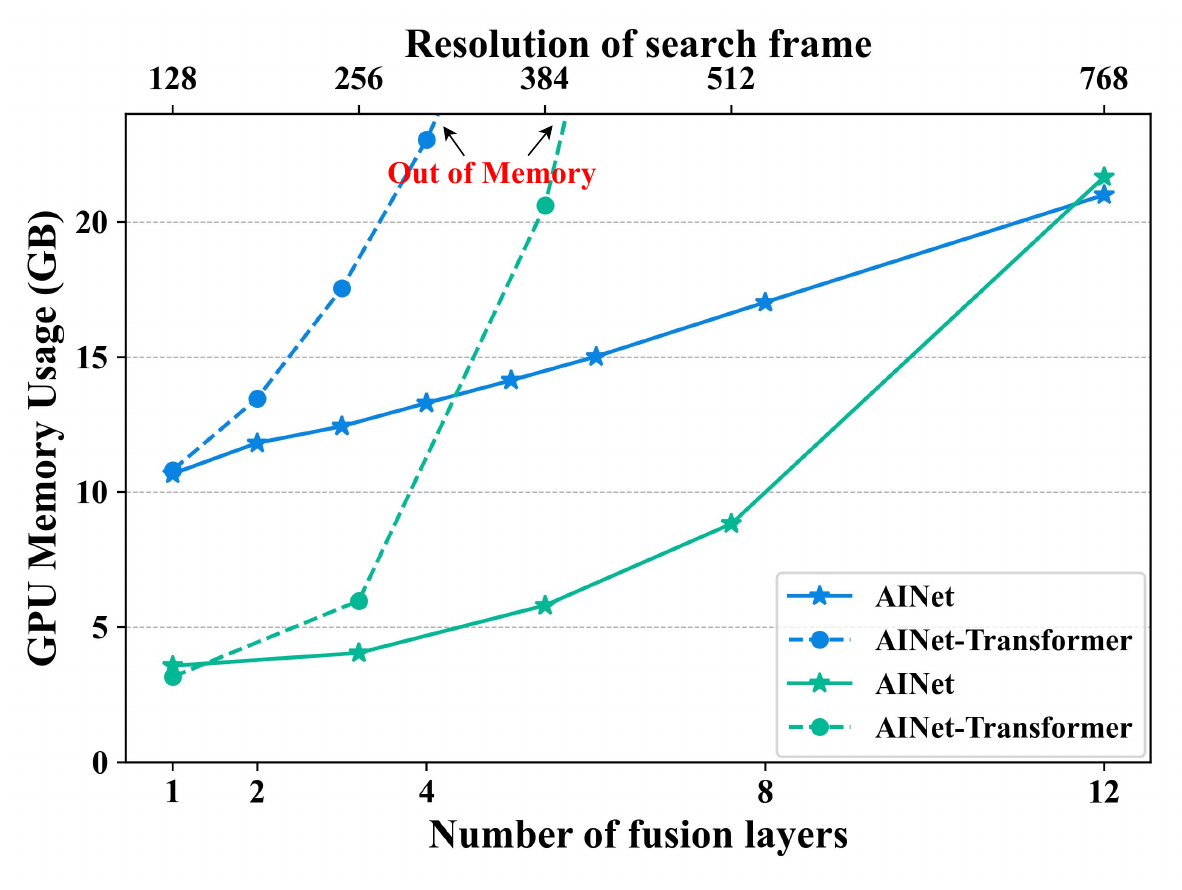}
     \caption{ Comparison of GPU memory usage between our proposed framework and a transformer-based approach under variations in layer count and resolution. The \textcolor{blue}{blue} line indicates resolution variation, and the \textcolor{c1}{green} line indicates variation in layer count.}
    \label{gpu}
\end{figure}

\noindent\textbf{Impact of different resolutions.}
Increasing the resolution of input images for performance improvement has been proven effective in many unimodal vision tasks~\cite{faster,yolov7,yoloworld}. However, existing RGBT tracking algorithms are constrained by the high computational complexity of the interaction module, preventing them from utilizing this strategy.  Benefiting from the computational efficiency of Mamba, AINet introduce a larger input resolution of $384\times384$ to further enhance performance. As shown in Table~\ref{ab_component}, AINet achieves new state-of-the-art performance with this resolution increase, demonstrating its effectiveness across multiple datasets. In addition, we compare the GPU memory usage of AINet and its Transformer variant (AINet-Transformer) with increased resolution Fig.~\ref{gpu}. In particular, we use cross-attention instead of DFM and self-attention instead of OFM. AINet-Transformer's memory usage grows quadratically with resolution, becoming impractical, while AINet scales linearly, maintaining efficiency. Note that the batch size is 1. These experiments fully demonstrate the efficiency advantage of AINet's interaction module.


\begin{table}[]
\centering
\caption{Ablation study of applying different number of layers on the LasHeR dataset.}
\setlength{\tabcolsep}{3.5mm}{
\renewcommand{\arraystretch}{1.0}
\resizebox{\linewidth}{!}{
\begin{tabular}{c|c|ccc} \toprule
Variants  & Layer index  & PR$\uparrow$ & NPR$\uparrow$ & SR$\uparrow$   \\ \hline
AINet-1 & 11   & 71.4 & 67.4 & 56.9 \\
AINet-3 & 0,6,11   & 71.9 & 68.1 & 57.4 \\
AINet-6 & 0,2,4,6,8,11   & 72.1 & 68.4 & 57.8 \\
AINet-12 & all   & \textbf{73.0} & \textbf{69.1} & \textbf{58.2}  \\ \bottomrule
\end{tabular}}}
\label{ab_layers}
\end{table}

\noindent\textbf{Impact of different layers.}
To verify the effectiveness of utilizing all layer features for RGBT tracking, we explore the impact of employing different numbers of layers. In Table~\ref{ab_layers}, we present three variants denoted as AINet-“n”, where “n” represents the number of layers applied. The results show that overall performance improves as the number of layers increases. Compared to using only the last layer for fusion, involving all layers leads to a performance increase of 1.6\%/1.7\%/1.3\% in PR/NPR/SR on LasHeR.
Furthermore, we analyze the resource constraints faced by existing interaction strategies when applying different numbers of layer features. As shown in Fig.~\ref{gpu}, the GPU memory usage of AINet-Transformer surges with additional layers, resulting in OOM (Out of Memory) at five layers. In contrast, our approach shows a linear correlation between computational resource requirements and the number of layers, allowing AINet to fully exploit information from all layers while balancing performance and efficiency.

\begin{table}[ht]
\centering
\caption{Comparison of performance, additional parameters, FLOPs, and FPS with advanced trackers on LasHeR. All algorithms' FPS are evaluated on a single RTX 4090 GPU.
}
\setlength{\tabcolsep}{2mm}{
\renewcommand{\arraystretch}{1.2}
\resizebox{\linewidth}{!}{
\begin{tabular}{c|c|cc|ccc}
    \toprule
    Methods & Pub. Info. & PR$\uparrow$ & SR$\uparrow$ & Params$\downarrow$ & FLOPs$\downarrow$ & FPS$\uparrow$ \\ \hline
    TBSI       & CVPR 2023       & 69.2 & 55.6 & 99.3M & 38.5G & 35.7 \\ \hline
    GMMT        & AAAI 2024     & 70.7 & 56.6 & 962.2M & 146.5G & 22.4 \\ \hline
    SDSTrack   & CVPR 2024       & 66.5 & 53.1 & \textbf{10.1M} & 13.1G & 18.4 \\ \hline
    Ours       & $-$       & \textbf{73.0} & \textbf{58.2} & 15.9M & \textbf{5.2G} & \textbf{38.1} \\ \bottomrule
\end{tabular}}}
\label{ab_efficiency}
\end{table}

\subsection{Efficiency Analysis} To verify the efficiency of our method, we perform a quantitative comparison with existing state-of-the-art methods. As shown in Table~\ref{ab_efficiency}, we present the number of parameters, computational burden compared to their respective baselines, and the inference speed of each model. Compared to the fully finetuned high-performance algorithms TBSI and GMMT, our approach is superior in all metrics. 
When compared to the partially finetuned advanced method SDSTrack, despite a slight difference in the number of parameters, our approach shows a significant advantage in performance and other efficiency metrics. Notably, while our approach excels in FLOPs metrics, it underperforms in inference speed. This is attributed to the fact that Mamba is not yet adapted to the available acceleration hardware.

\section{Conclusion}
In this paper, we explore for the first time the potential of Mamba in RGBT tracking by designing a novel All-layer Interactive Network (AINet), which effectively integrates information from all layers to achieve robust tracking performance.
The core concept of AINet is to introduce a progressive fusion of Mamba to enable efficient and effective all-layer modality interactions. Specifically, AINet designs a difference-based fusion Mamba that enhances modality interactions at each layer of the backbone network by modeling modality differences. Additionally, an order-dynamic fusion Mamba is designed to perform interactions across all layer features, mitigating the risk of early token information loss.
Extensive experiments demonstrate the superior performance of the proposed method in four popular RGBT tracking benchmarks.
Currently, we still adopt ViT as the backbone network for modal feature extraction, thus there is room for improvement in the overall model efficiency. To enhance efficiency, we plan to adopt Mamba as the backbone network in the future. Nevertheless, there is no available Mamba-based tracking pre-training model, thus we intend to establish a model distillation method to create a pure Mamba-based multimodal tracking network. 







\bibliographystyle{IEEEtran}
\bibliography{AINet}

\end{document}